\documentclass[conference]{IEEEtran}
\ifCLASSINFOpdf
\else
\fi
\usepackage{amsmath}
\usepackage[linesnumbered,ruled,vlined]{algorithm2e}

\usepackage[flushleft]{threeparttable}
\usepackage{graphicx}

\usepackage{amsfonts}
\usepackage{etoolbox}
\makeatletter
\patchcmd{\@makecaption}
  {\scshape}
  {}
  {}
  {}
\makeatother

\newcommand{\fatx}{\mathbf{x}}
\newcommand{\fatz}{\mathbf{z}}
\newcommand{\fatxo}{\mathbf{x}^{(1)}}
\newcommand{\fatzo}{\mathbf{z}^{(1)}}
\newcommand{\fatxn}{\mathbf{x}^{(n)}}
\newcommand{\fatzn}{\mathbf{z}^{(n)}}
\newcommand{\fatxm}{\mathbf{x}^{(m)}}
\newcommand{\fatzm}{\mathbf{z}^{(m)}}
\newcommand{\fatxi}{\mathbf{x}^{(i)}}
\newcommand{\fatzi}{\mathbf{z}^{(i)}}
\newcommand{\fatxj}{\mathbf{x}^{(j)}}
\newcommand{\fatzj}{\mathbf{z}^{(j)}}
\newcommand{\fatxb}{\bar{\mathbf{x}}}
\newcommand{\fatxbo}{\bar{\mathbf{x}}^{(1)}}
\newcommand{\fatxbi}{\bar{\mathbf{x}}^{(i)}}

\newcommand{\fatxbn}{\bar{\mathbf{x}}^{(n)}}
\newcommand{\fatn}{\mathbf{n}}

\newcommand{\qth}{q_\theta}
\newcommand{\pth}{p_\theta}
\newtheorem{lemma}{Lemma}
\newcommand{\LL}{\mathcal{L}}

\DeclareMathOperator*{\argmax}{arg\,max}
\DeclareMathOperator*{\argmin}{arg\,min}
\begin{document}
%
\title{Unsupervised feature learning with discriminative encoder}

\author{\IEEEauthorblockN{Gaurav Pandey and Ambedkar Dukkipati}
\IEEEauthorblockA{Department of Computer Science and Engineering \\
Indian Institute of Science\\
Email: \textit{\{gauravp, ambedkar\}@iisc.ac.in}
}}


%


\maketitle

\begin{abstract}
In recent years, deep discriminative models have achieved extraordinary performance on supervised learning tasks, significantly outperforming their generative counterparts. However, their success relies on the presence of a large amount of labeled data. How can one use the same discriminative models for learning useful features in the absence of labels? We address this question in this paper, by jointly modeling the distribution of data and latent features in a manner that explicitly assigns zero probability to unobserved data. Rather than maximizing the marginal probability of observed data, we maximize the joint probability of the data and the latent features using a two step EM-like procedure. To prevent the model from overfitting to our initial selection of latent features, we use adversarial regularization. Depending on the task, we allow the latent features to be one-hot or real-valued vectors, and define a suitable prior on the features. For instance, one-hot features correspond to class labels, and are directly used for unsupervised and semi-supervised classification task, whereas real-valued feature vectors are fed as input to simple classifiers for auxiliary supervised discrimination tasks. The proposed model, which we dub dicriminative encoder (or DisCoder), is flexible in the type of latent features that it can capture. The proposed model achieves state-of-the-art performance on several challenging tasks. 
\end{abstract}


%
\IEEEpeerreviewmaketitle

\section{Introduction}
Deep neural networks have achieved extraordinary performance for several challenging supervised learning tasks in recent years. Convolutional neural networks~\cite{fukushima1982neocognitron, le1990handwritten} have greatly improved the state-of-the-art for several problems in computer vision including classification~\cite{jia2014caffe}, object detection~\cite{girshick2015fast}, semantic segmentation~\cite{long2015fully} etc. Similarly, recurrent neural networks (and its variants)~\cite{hochreiter1997long} have greatly improved the state-of-the-art on several sequential modeling tasks, such as speech-to-text generation~\cite{graves2013speech}, language translation~\cite{sutskever2014sequence} etc.

However, access to large amount of labeled data has been crucial for the success of neural networks for most of the above tasks. Since obtaining labeled data in such large quantities is expensive, one may wonder if it is possible to make use of unlabeled data for learning meaningful representations. This question has been addressed by several researchers using a plethora of techniques. 

Among these models, generative approaches model the joint distribution of the visible and the latent features. The marginal log-probability of the observed data is then maximized. Examples of such models in deep learning include restricted Boltzmann machines~\cite{hinton2002training}, variational autoencoders~\cite{kingma2013auto, rezende2014stochastic} and generative adversarial networks~\cite{goodfellow2014generative, donahue2016adversarial}. Variants of these models have been successfully used for learning categorical embeddings for unsupervised and semi-supervised classification tasks~\cite{springenberg2015unsupervised, salimans2016improved, kingma2013auto}.

Another set of models use self-supervision for learning real valued embeddings of data~\cite{noroozi2016unsupervised, doersch2015unsupervised, wang2015unsupervised}. These embeddings are then fed to a much smaller network for solving an auxiliary classification problem. The success of these models relies on the assumption that networks trained using self-supervision will learn meaningful embeddings that generalize well for other tasks. 

In this paper, we propose a general approach for learning latent representations from unlabeled data. The model, which we refer to as discriminative encoder (or DisCoder) is completely specified by the encoding and the prior distribution on the latent features. The model makes no assumptions about the distribution of the latent features. However, in our experiments we limit ourselves to categorical and normally distributed latent features only, and show that these latent features consistently achieve the state-of-the-art performance on several tasks. 

The rest of the paper is organized as follows: In Section~\ref{sec:model}, we motivate the choice of our model, and discuss the steps involved in training the model. We also discuss an adversarial regularization strategy that prevents the model from getting stuck to its initial configuration. In Section~\ref{related}, we discuss other models that are used for unsupervised representation learning. In Section~\ref{experiments}, we present the results achieved by our model on several tasks including clustering, semi-supervised and auxiliary classification.

\section{The proposed model}~\label{sec:model}
In this section, we will motivate our choice of the model for unsupervised feature learning. We will discuss optimization strategies for the model that utilize minibatches of data, and are nearly as fast as models used in supervised learning. The observed features will be denoted by $\fatx = (x_1,\ldots, x_d)$, whereas the corresponding latent features will be denoted by $\fatz = (z_1, \ldots, z_m)$. In particular bold font will be used for vectors, while normal font will be used to denote the corresponding components. The distribution $\pth(\fatz|\fatx)$ will be referred to as the encoding distribution, whereas $\pth(\fatx|\fatz)$ will be referred to as the decoding distribution. The parameters of the networks will be denote by $\theta$. The output of the encoding network will be denoted as $\phi$. We will denote that the number of samples in the training data by $n$, and the size of the latent features by $K$.

\subsection{Motivation}
A common approach to address the problem of unsupervised learning with latent features is by defining a joint distribution over the observed and the latent features. Such a model will be referred to as generative model in the sequel. One can then maximize the marginal distribution of the observed features. 
\begin{equation}
\pth(\fatx) = \sum_{\fatz} \pth(\fatx, \fatz)
\end{equation}
The MAP estimate of the latent features given the observed features can then be used as a representation for the data.
\begin{equation}
\hat{\fatz} =  \argmax_{\fatz} \log \pth(\fatx, \fatz)
\end{equation}

The above approach is one of the most common approaches for modeling latent features, and has been successfully used in Gaussian mixture models (GMM), hidden Markov models (HMM), restricted Boltzmann machines (RBM), variational autoencoder (VAE), latent Dirichlet allocation (LDA) etc. Despite its almost ubiquitous success, this approach for learning latent features suffers from a fatal flaw: It is theoretically  possible to maximize the $\pth(\fatx)$ without changing the encoding distribution $\pth(\fatz|\fatx)$ at all. This is particularly true when the choice of decoding distribution $\pth(\fatx|\fatz)$ is flexible enough to fit the data distribution. In such a scenario, the model can choose to ignore the latent features $\fatz$ in the decoding distribution $\pth(\fatx|\fatz)$ by equating $\pth(\fatx|\fatz)$ to $\pth(\fatx)$. When this happens,
\begin{equation}
\pth(\fatz|\fatx) = \frac{\pth(\fatx| \fatz)\pth(\fatz)}{\pth(\fatx)} = \frac{\pth(\fatx)\pth(\fatz)}{\pth(\fatx)} = \pth(\fatz)\,,
\end{equation}
that is, the latent features $\fatz$ are completely independent of the observed features $\fatx$.

In general, the choice of decoding model is simple enough to prevent this from happening. For instance, in HMM, RBM, VAE and LDA, the decoding distribution factorizes completely, whereas in GMM, the decoding distribution is a Gaussian distribution. This forces the model to utilize the latent features to effectively model the data distribution. 

Hence, generative models rely on weak decoding distributions for learning useful representations. In general, a good generative model doesn't guarantee a useful latent representation. The problem lies in the fact, that generative models spend most of their time optimizing $\pth(\fatx)$, and may or may not choose to care about the encoding distribution $\pth(\fatz|\fatx)$. To address this specific issue, we define a model that specifically assigns zero probability to any point $\fatx$ that doesn't occur in the training data. One can see that such a modeling strategy is intinsically used by discriminative models for supervised learning tasks. In particular, if $(\fatx^{(1)}, \fatz^{(1)}),\ldots, (\fatx^{(n)}, \fatz^{(n)})$ are the observed points and their labels, the joint log-likelihood $\mathcal{L}$ can be written as
\begin{equation}
\mathcal{L} = \sum_{i=1}^n \log \pth(\fatx^{(i)}) + \sum_{i=1}^n \log \pth(\fatz^{(i)}|\fatx^{(i)})
\end{equation}
In discriminative models, the first term in the above equation is completely independent of the parameters in the second term, and hence can be maximized independently. The maximum occurs, when $\pth(\fatx)=\frac{1}{n}$ for all the observed points, and $0$ for any point that doesn't occur in the training data. 

Unfortunately, when the labels $\fatz$ are unknown, the same strategy can't be used, since the model collapses to a single $\fatz$ with $\pth(\fatz|\fatx)=1$ for all $\fatx$. To prevent this from happening, we couple the encoding distribution $\pth(\fatz|\fatx)$ with a function $f(\fatz)$, and define a new distribution
\begin{equation} \label{qdef}
\qth(\fatx, \fatz) = {\pth(\fatz|\fatx)f(\fatz)}
\end{equation}
To obtain the value of $f$ at $\fatz$, we force the distribution to satisfy the following constraints:
\begin{enumerate}
\item The marginal distribution of $\qth(\fatx, \fatz)$ over $\fatz$ is $\pth(\fatz)$. This is done to prevent the model from collapsing to a single $\fatz$. This also allows us to incorporate the prior information about the latent features into our model. For instance, if we know that the components of $\fatz$ should form a Markov chain, we can incorporate that information easily into the model.
\item The distribution $\qth$ assigns $0$ probability to any point that doesn't occur in the training data. This is done to prevent the model from spending its time and effort in training $\qth(\fatx)$. In the words of Vapnik, ``One should solve the problem directly and never solve a more general problem as an intermediate step".
\end{enumerate}
\begin{lemma}
Let $\qth$ be a distribution of the form given in \eqref{qdef} that satisfies the constraints above. Then
\begin{equation}
\qth(\fatx,\fatz) = \frac{\pth(\fatz|\fatx)\pth(\fatz)}{\sum_{j=1}^n \pth(\fatz|\fatxj)}\,,
\end{equation}
for any $\fatx$ observed in the training data and, $0$ otherwise.
\end{lemma}
This is the distribution that we will optimize for unsupervised feature learning in the rest of the paper. \textit{Note that the model is completely specified by the choice of the prior $\pth(\fatz)$ and the encoding distribution $\pth(\fatz|\fatx)$.}

\subsection{Discriminative Encoder (DisCoder)}
In the previous section, we motivated the choice of the model used for unsupervised feature learning in this paper. The model will be referred to as discriminative encoder (or DisCoder) in the sequel. In this section, we will discuss the model in further detail, and will see, why the name \textit{discriminative encoder} is apt for this model.

Given an $i.i.d$ sequence of unlabeled samples $\fatxo, \ldots, \fatxn$, the joint likelihood function can be written as a function of the corresponding latent features $\fatzo, \ldots, \fatzn$, and parameters of the model $\theta$.
\begin{equation}
\mathcal{L}(\theta, \fatzo, \ldots, \fatzn) = \sum_{i=1}^n \log q_\theta(\fatxi, \fatzi)\,,
\end{equation}
where $\qth(\fatx, \fatz)$ is as defined in~\eqref{qdef}. In order to optimize the above objective, we use an EM-like procedure, that alternates between the selection of $\fatzi$ and optimization with respect to $\theta$.

In the selection step, for each $\fatxi$ in the training data, we find the best $\fatz$, that is, the $\fatz$ that maximizes $\log \qth(\fatxi,\fatz)$. The quantity $\log \qth(\fatxi, \fatz)$ can be expanded as
\begin{equation} \label{eq:latent}
\log q_\theta(\fatxi, \fatz) = \log p_\theta(\fatz|\fatxi) + \log p_\theta(\fatz) - \log \sum_{j=1}^n p_\theta(\fatz|\fatxj)
\end{equation}
Hence, the optimal latent representation  should be such it maximizes $\log \pth(\fatz|\fatxi)$, but minimize $\log \sum_{j=1}^n \pth(\fatz|\fatxj)$. In other words, the latent representation should have high probability of being assigned to $\fatxi$, but low probability of being assigned to any other $\fatxj, j\neq i$. 

Once we have found the optimal latent representation for $\fatxi$, we equate it to $\fatzi$. Next, we optimize the objective with respect to the parameters of the model. In expanded form, the objective as a function of $\theta$ can be written as
\begin{equation} \label{liketheta}
\mathcal{L}(\theta) = \sum_{i=1}^n \log \pth (\fatzi|\fatxi) - \sum_{i=1}^n \log \sum_{j=1}^n  \pth (\fatzi|\fatxj)
\end{equation}
For instance, when $\pth(\fatz|\fatx)$ is normally distributed with mean $\phi(\fatx)$ and variance $1/2$, the objective can be written as
\begin{align}~\label{latent_optimize}
-\mathcal{L}(\theta) = &\sum_{i=1}^n ||\phi(\fatxi)-\fatzi||^2 \notag\\
	&+ \sum_{i=1}^n \log \sum_{j=1}^n \exp(-||\phi(\fatxj)-\fatzi||^2)
\end{align}
Since $\fatzi$ are fixed for this step, the terms that depend on $\fatzi$ alone, have been removed from the objective. For a fixed $\fatxi$, this step trains the network to maximize the probability of the $\fatzi$ selected in the previous step while simultaneously lowering the probability of other $\fatzj, j \neq i$. 

Both the steps of training try to ensure that the learnt representations are as dissimilar to each other as possible, while simultaneously satisfying the requirement of prior distribution. Hence, the representations learn to capture those features that vary among the samples, while completely ignoring the features common to all the samples. Hence, we name the model as discriminative encoder (or DisCoder). \\

\noindent\textbf{Semi-supervised learning:} If we have access to a set of labelled samples, DisCoder can utilize those samples to improve upon the learnt embeddings. We achieve this by adding a term to the objective to maximize the conditional log-likelihood over the labelled samples. If we denote the set of labelled samples as $\{(\fatxo_s, \fatzo_s),\ldots, (\fatxm, \fatzm)\}$, the new objective can be written as:
\begin{align}
\LL&(\theta, \fatzo, \ldots, \fatzn) \\
&= \sum_{i=1}^n \log \qth(\fatxi, \fatzi) + \sum_{i=1}^m \log \pth(\fatzi_s|\fatxi_s)\,,
\end{align}

\subsection{Optimization}
As mentioned in the previous section, the optimization proceeds in the following two steps:
\begin{enumerate}
\item Latent feature selection
\item Encoding network optimization
\end{enumerate}
The latent feature selection step differs depending on the choice of the latent features, while the encoding network optimization step remains essentially the same irrespective of the latent features. We will first discuss the latent feature selection step for $2$ special cases and then discuss the network optimization step. \\

\noindent\textbf{Latent feature selection:} While the proposed model is general enough to handle any encoding distribution, in this paper, we restrict ourselves to categorical and Gaussian distribution. Categorical distribution is used, when the latent features are one-hot vectors. For a fixed $\fatxi$, the mean of the encoding distribution $\phi(\fatxi)$ is the output of the network. The latent feature selection step for categorically distributed latent features can be obtained by exponentiating equation~\eqref{eq:latent}
\begin{align}~\label{eq:latent_cat}
\fatzi = \argmax_{\fatz} \frac{\prod_{k=1}^K \phi_k(\fatxi)^{z_k}}{\sum_{j=1}^n\prod_{k=1}^K \phi_k(\fatxj)^{z_k}}\end{align}
In particular, $z_l^{(i)}=1$, if 
\begin{equation}
l = \argmax_{k \in \{1,\ldots, K\}} \frac{\phi_k(\fatxi)}{\sum_{j=1}^n \phi_k(\fatxj)}
\end{equation}
The last equation follows from the fact that only one component of $\fatz$ can be non-zero at a time. It can be computed efficiently and independently for each component.

For real valued latent features $\fatz$, we assume that the prior over $\fatz$ follows a uniform distribution over a sphere, while the encoding distributed is normally distributed with a constant variance $\lambda$, which is selected by cross-validation on a validation set for a secondary task. For a fixed $\fatxi$, the mean of the encoding distribution $\phi(\fatxi)$ is the output of the network. The latent feature selection step for real valued vectors can be written as:
\begin{align} \label{eq:latent_normal}
\fatzi = &\argmin_{\fatz} \frac{||\phi(\fatxi)-\fatz||^2}{2\lambda}   \notag\\
& + \log \sum_{j=1}^n \exp\left( - \frac{||\phi(\fatxj)-\fatz||^2}{2\lambda}\right)
\end{align}
The above objective is a differentiable function of $\fatz$. Hence, we use stochastic gradient descent (SGD) for optimization. Moreover, we optimize the restriction of the third term in the objective over a small batch of $1000$ samples only. In particular, we keep a queue of $1000$ samples on which the model was trained recently, and use the samples in this queue for minimizing the objective.

\noindent\textbf{Encoding network optimization:} For fixed $\fatx$ and $\fatz$, the encoding distribution $\pth(\fatz|\fatx)$ is a differentiable function of $\theta$. Hence, we use minibatch SGD to maximize the objective $\LL(\theta)$ defined in~\eqref{liketheta}. Given a minibatch of samples, we optimize the restriction of the objective to this minibatch, while ignoring the examples from other minibatches. In particular, the summation inside the log of second term in \eqref{liketheta} is evaluated over the minibatch only. This greatly reduces the computational complexity of optimization from $\mathcal{O}(n^2)$ to $\mathcal{O}(nb)$, where $b$ is the batchSize. Furthermore, the randomness introduced by the restriction serves to regularize the encoding network, and encourages it to explore other choices of $\fatz$.

\subsection{Regularizing the model}~\label{regularization}
As mentioned in the previous section, the training of the model alternates between latent feature selection and encoding network optimization. The latent features selected at the beginning of training are completely random.  Now, if the encoding network is sufficiently powerful, and the encoding network optimization step proceeds for a sufficiently long time, the encoding network may attempt to fit to the initial features, despite the fact that they are completely random.

This is a problem faced by almost all the approaches for unsupervised representation learning. For instance, in clustering based approaches~\cite{yang2016joint, xie2016unsupervised, springenberg2015unsupervised}, if the encoding network that maps the samples to clusters is sufficiently powerful, and the training of encoding network proceeds for a sufficiently long time, keeping the clusters fixed, any choice of the clusters can be predicted with absolute certainty. Even in the case of autoencoders, if the encoding and the decoding networks are sufficiently powerful, the autoencoder will be able to map any image to any representation, while the decoding network will be able to map the same representation back to the image. (In fact, in our experiments on 2D-representations of MNIST digits, we found that if the encoding and decoding networks are deep enough, the representations learnt by an autoencoder don't change much during the course of training.)

In general, a number of reasons prevent an unsupervised learning algorithm from getting stuck in the initial choice of representations:
\begin{enumerate}
\item The encoding network is often optimized for a single stochastic gradient descent update, and is immediately followed by a secondary step, such as clustering (or latent feature selection in our case).
\item The encoding network is regularized heavily to force it to learn simpler mappings.
\item Stochastic gradient descent training in neural networks prefers simple mappings that result in flat minima~\cite{keskar2016large}.
\end{enumerate}
 
Badly selected latent features can be escaped easily, if the variance of the encoding distribution is high. This can be observed from equation~\eqref{eq:latent_normal} for normally distributed latent features. When $\lambda$ is high, a slight change in the latent representation $\fatz$, doesn't change the objective by much. Hence, the stochasticity introduced by the optimization technique is often enough for the model to explore other choices for $\fatz$. However, as $\lambda$ gets smaller, the expression gets more and more sharply trenched around $\phi(\fatxi)$. This makes it almost impossible for the model to assign any value other than $\phi(\fatxi)$ to $\fatz$, despite the stochasticity of training.

Hence, primarily for the above reason, we fix the variance of the encoding distribution to a constant, which is selected via cross-validation on a validation set for a secondary task, when the latent features are normally distributed. This approach works, since the latent space is continuous, and several choice of latent features have high probability under the encoding distribution. However, if we try to achieve the same for categorically distributed latent features, we end up with distributions that are almost uniform over all the labels. The DisCoder ends up learning nothing meaningful. The same is true for any choice of discrete distribution, since the latent features are far apart. To prevent this from happening, we use adversarial regularization for categorical latent features, which is discussed next.

\noindent \textbf{Adversarial regularization:} The adversarial regularization works as follows: While training the DisCoder to maximize $\sum_{i=1}^n \log \qth(\fatxi, \fatzi)$ for real samples $\fatxo, \ldots, \fatxn$, we simultaneously train it to be totally confused about a set of \textit{fake} samples $\fatxbo,\ldots, \fatxbn$. We force this by maximizing the probability over all the latent features for fake images. Hence, the new objective becomes:
\begin{equation}
\mathcal{L}(\theta) = \sum_{i=1}^n \log \qth(\fatxi, \fatzi) + \sum_{i=1}^n \sum_{\fatz}p(\fatz) \log\pth(\fatz|\fatxbi)
\end{equation}
The above regularization forces the model to learn mappings from data to categories that are completely unsure about samples that don't occur in the training data. A similar scheme for regularization was also used in~\cite{springenberg2015unsupervised}.

\noindent\textit{Note: The adversarial regularization used in this paper must not be confused with adversarial training used in~\cite{goodfellow2014explaining}. We choose the term adversarial regularization for lack of a better name.}

In order to complete the description of the model, we also need a way to generate fake samples. We use a generator to generate fake images, while we use negative sampling to generate fake documents.

\noindent \textbf{Class-conditioned generation:}  In order to generate fake image samples, one can couple the encoder with a generator $G$, that takes Gaussian noise $\fatn$ and latent features $\fatz$ as input and generates fake samples $\fatxb$ as output, that is,
$$\fatxb = G(\fatn, \fatz)$$
The generator is trained to generate samples $\fatxb$, such that the encoder can extract $\fatz$ back from $\fatxb$. We achieve this by training the generator to maximize $\log \pth(\fatz|\fatxb = G(\fatn, \fatz))$. Note that the discriminator is simultaneously being trained to be completely confused about the latent features of the fake examples. This creates an adversarial game akin to the adversarial game of generative adversarial networks~\cite{goodfellow2014generative}. The choice of the generative model allows us to visualize the latent features, and hence, determine if the model is learning anything meaningful. In particular, if we wish to determine the information captured by the $i^{th}$ latent feature, we set it to $1$ and the other latent features to $0$, couple it with Gaussian noise and pass it through the generator. 

\noindent \textbf{Feature matching:} Class-conditioned generation results in images that are sharp with the object immediately identifiable. Note that the encoding network is simultaneously being trained to be totally confused about the generated samples. However, if the underlying class of the fake image is immediately identifiable, training a network to be confused about the image results in unstable training. Hence, while class conditioned generation is useful when the underlying task is to generate sharp images, it is not really suitable for classification tasks. 

In order to address this problem, we use feature matching for classification tasks. Noise is passed through the generator to generate a batch of fake images $\fatxb$, that is, $\fatxb = G(\fatn)$. The statistics of the fake batch is then compared with the statistics of the real batch. As suggested in~\cite{salimans2016improved}, we pass the fake batch as well as the real batch through the encoding network, and compute the squared difference between the means of encodings of the real and fake batch for the penultimate layer. The generator is trained to minimize the distance between the average encodings for the penultimate layer. The reultant images aren't very sharp. However, this approach works quite well for classification task as has been observed in~\cite{salimans2016improved}.

\noindent \textbf{Negative Sampling:} To generate fake documents, we randomly select words from the vocabulary, based on the frequency with which they occur in the corpus. The size of the document is Poisson distributed whose mean is the mean of the length of the documents that occur in the corpus

Algorithm~\ref{algo:b} combines all the above steps for training the model.

\subsection{Complexity of training}
As mentioned in the previous sections, the complexity of each step (latent feature selection as well as network optimization) is quadratic in the size of the batch, that is, $\mathcal{O}(b^2)$. However, when the batch size is small enough ($< 2000$ samples), the empirical running time depends linearly on the batch size rather than quadratically. This is because, most of the running time is spent in forward and backpropagating through the network, which needs to be done only once for the entire batch). Very little time is utilized in the computation of the distribution of the Discoder or its gradient from the output of the network. Empirically, we observed that DisCoder takes approximately the same time as Improved GAN~\cite{salimans2016improved} for semi-supervised learning tasks.

\begin{algorithm}
\KwIn{Observed features $\fatxo,\ldots, \fatxn$}
\KwOut{An encoder $\pth(\fatz|\fatx)$ and generator $G$ (if adversarial regularization is enabled) } 
\vspace{.2cm}
Repeat the following steps until convergence 
\begin{enumerate}
\item Randomly select a batch of observed features $B$.
\item Forward propagate them through the encoder to obtain the encoding distribution $\pth(\fatz|\fatx), \fatx \in B$.
\item Select the latent features $\fatz_\fatx$ for all $\fatx \in B$ by maximizing 
$\log \qth(\fatx, \fatz) $ for each $\fatx$ in the batch.
\item Train the encoding network to maximize
$$\LL(\theta) = \sum_{\fatx \in B}\log q_{\theta} (\fatx, \fatz_\fatx)$$
\item If semi-supervised learning is enabled:
	\begin{enumerate}
	\item Randomly select a batch $B_s$ of labeled samples.
	\item Train the encoding network to maximize $\sum_{(\fatx, \fatz) \in B_s} \log \pth(\fatz|\fatx)$.
	\end{enumerate}
\item If class conditioned generation is enabled:
	\begin{enumerate}
	\item Concatenate noise $\fatn$ with the latent features obtained in the previous step $\fatz_\fatx$ and propagate them through the generator
	to get a batch of fake samples $\bar{B}$.
	\item Train the encoding network to maximize 
	$$\sum_{\fatx \in \bar{B}} \sum_{\fatz}p(\fatz) \log\pth(\fatz|\fatxbi)$$
	\item Train the generator to maximize $\log \pth(\fatz|\fatxb = G(\fatn, \fatz))$
	\end{enumerate}
\item If feature matching is enabled:
	\begin{enumerate}
	\item Forward propagate noise $\fatn$ through the generator to generate a batch of fake samples.
	\item Train the encoding network to maximize
	$$\sum_{\fatx \in \bar{B}} \sum_{\fatz}p(\fatz) \log\pth(\fatz|\fatxbi)$$
	\item Let the mean of the penultimate layer of the encoder for the fake batch be $\phi_{\bar{B}}$ and for the real batch be $\phi_B$. 
	\item Train the generator to minimize the squared distance between the average encodings for the real and fake batch for the penultimate layer.
	\end{enumerate}	
\item If negative sampling is enabled:
	\begin{enumerate}
	\item Create a batch of fake documents $\bar{B}$ by sampling words in the vocabulary based on their frequency with which they occur in the corpus.
	\item Train the encoding network to maximize  
	$$\sum_{\fatx \in \bar{B}} \sum_{\fatz}p(\fatz) \log\pth(\fatz|\fatxbi)$$
	\end{enumerate}
\end{enumerate}
\caption{Minibatch training of DisCoder}
\label{algo:b}
\end{algorithm}

\section{Related works}~\label{related}
Unsupervised learning is a well studied problem in machine learning, and a plethora of techniques exist for addressing this problem, including probabilistic and non-probabilistic approaches. Most of the probabilistic models for unsupervised learning are generative models, that is, they explicitly model the probability of the observed data. In this paper, we will restrict ourselves to discuss only those models that are employed in deep learning.

\subsection{Generative models} 
Representation learning in generative models proceeds by defining a joint distribution over the visible features and the latent features. One can then optimize the marginal log-likelihood (or its approximation) over the observed data. Examples of such models include Boltzmann machines (RBM~\cite{hinton2002training}, DBM~\cite{salakhutdinov2009deep}),  variational autoencoders (VAE)~\cite{kingma2013auto,rezende2014stochastic} and generative adversarial networks (GAN)~\cite{goodfellow2014generative}. 

RBM and DBM have proved to be excellent for unsupervised learning on simple datasets such as MNIST. However, training as well as inference in deep extensions of these models is quite challenging, especially when convolutional layers are employed in these models. This severely limits the application of these models for unsupervised feature learning. 

In contrast, variational autoenoders~\cite{kingma2013auto, rezende2014stochastic} and its deterministic variant~\cite{bengio2007greedy} utilize deep neural networks for modeling, and are comparatively simpler to train. These models learn embeddings so as to minimize the reconstruction error. Variational autoencoders have been successfully used for a wide variety of tasks in deep learning, that include image captioning~\cite{xu2015show}, semi-supervised learning~\cite{kingma2014semi} etc.

Among the generative models, the model most relevant to our work is GAN~\cite{goodfellow2014generative}. In a GAN, a discriminator is trained to distinguish between fake and real samples, while a generator is simultaneously trained to generate images from a latent distribution that fool the discriminator. Recent works involving GANs~\cite{donahue2016adversarial, dumoulin2016adversarially} have explored methods for inferring the latent representations from the samples, and successfully utilized these latent representations for several auxiliary tasks. 

Variants of GANs have also been used for unsupervised and semi-supervised learning tasks. In particular, CatGANs~\cite{springenberg2015unsupervised} learn a categorical latent representation for the data by maximizing the mutual information between the data and the categorical labels, while simultaneously maximizing the entropy over the class labels for fake images. Adversarial autoencoders~\cite{makhzani2015adversarial} learn a mapping from the data to the latent space so as to minimize reconstruction error. However, unlike variational autoencoders, they use a GAN framework to enforce the prior on the latent features. This allows the model to learn hybrid (discrete + continuous) embeddings, which isn't possible for variational autoencoders.

\subsection{Self-supervised learning:} 
Self-supervised learning proceeds by generating labels from the data using information present in the structure of the data. There has been a growing interest in such methods, since no extra effort is need for labeling these samples. For instance, Dosovitskiy \textit{et al.}~\cite{dosovitskiy2014discriminative} train a model for training a convolutional network that assigns each image to its own class. In particular, each image is used as a seed to generate a class of images by applying various transformations. Hence, the output layer of the  CNN grows linearly with the number of images in the dataset, severely limiting the scalability of the model.

Another set of methods force the model to learn the relative position of various patches in the image with respect to each other. For instance, jigsaw networks~\cite{noroozi2016unsupervised} permute the patches of the image, and train the network to predict the correct permutation. Similarly, context prediction networks~\cite{doersch2015unsupervised} are trained to predict the correct relationship relationship between two patches. To successfully complete this task, the models are forced to learn representations that capture the global structure of the image. 
 
Models that rely on videos as training data, attempt to learn features by exploiting the motion information present in videos. In particular, the approach in~\cite{doersch2015unsupervised}, identifies two patches that correspond to the same trajectory in a video, and minimizes the distance between their representations. This forces the model to learn feature that are invariant to orientation of the object in the video.

\section{Experiments}~\label{experiments}
In order to evaluate the capability of the model for learning meaningful representations, we evaluate the learnt representations across several tasks. For each task, we evaluate our model against the state-of-the-art models for that task. The code for the experiments in this paper is available at \textit{https://github.com/gauravpandeyamu/DisCoder}.

\subsection{Unsupervised and semi-supervised classification}
We evaluate the one-hot embeddings learnt by the model for the task of clustering on MNIST, and 20-newsgroup dataset. We also evaluate the embeddings for the task of semi-supervised classification on CIFAR-10 dataset.  In all our experiments, we use a batch size of 100 images, and Adam optimizer with a constant learning rate of $0.0002$. Since neural networks are likely to get stuck in local optima, we repeat the experiment $10$ times, and report the mean and the standard deviation.\\

\noindent\textbf{MNIST:} The MNIST~\cite{lecun1998gradient} is a relatively simple dataset of digits from $0$ to $9$. We normalize the images, by dividing the pixel intensities by 255. We evaluate DisCoder on this dataset for the clustering task into 20 clusters. We use generative adversarial regularization for regularizing the DisCoder as discussed in Section~\ref{regularization}. The architecture of the generator and encoding network are provided in Table~\ref{tab:MNIST_arch}. In order to evaluate the accuracy of the clustering algorithm, we need to associate each cluster with a label. Towards that end, we compute the intersection of the cluster with all the classes. The label of the class with the maximum intersection, is assigned to the cluster.

The clustering error achieved by the various models on MNIST dataset is shown in Table~\ref{tab:MNIST_res}. 
The accuracy is computed as follows: For each cluster $i$, we identify the sample $\fatx$ that maximizes $p(\fatz=i|\fatx)$. Next, the label of the selected $\fatx$ is assigned to all the samples in this cluster. The test error is computed based on the labels assigned to the points using this procedure. Such a scheme is also used for evaluating clustering in~\cite{makhzani2015adversarial}.

For completeness, we also show the results, when adversarial regularization isn't used. As can be observed, adversarial regularization is extremely crucial for achieving good performance, when one-hot embeddings are used in DisCoder. Also noteworthy, is the high standard deviation of an unregularized DisCoder, which indicates that the model is overfitting to the initial assignment.

\begin{figure}[!t]
\centering
\includegraphics[width=2.8in]{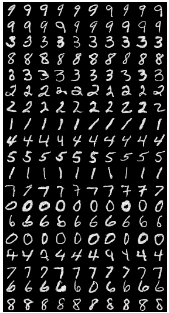}
\caption{Generated samples from the unsupervised MNIST generator conditioned on $\fatz$. Each row corresponds to a different component of $\fatz$ set to 1. Note that despite the fact that training was done in an unsupervised fashion, the generator was successfully able to associate different clusters with different classes. }
\label{fig_MNIST}
\end{figure}

The DisCoder achieves the state-of-the-art results for clustering on MNIST, significantly outperforming other methods. Since the generator is trained along with the encoder, we can visualize the images associated with each latent feature by setting the corresponding component of $\fatz$ to $1$, concatenating it with noise and passing it through the generator. The resultant images are shown in Figure~\ref{fig_MNIST}. As expected, each latent feature has learned to associate itself with images of a single class.

\begin{table}
\centering
\renewcommand{\arraystretch}{1.3}
\begin{threeparttable}

\caption{Network architecture for MNIST}
\label{tab:MNIST_arch}
\begin{tabular}{c c }
\hline
Generator & Encoder\\
\hline
256 ConvT 4x4, relu  & 64 Conv 4x4, stride 2, leaky relu\\
128 ConvT 3x3, stride 2, bn, relu  & 128 Conv 4x4, stride 2, bn, lrelu\\
 -----  & 128 Conv 3x3, stride 1, bn, lrelu \\
64 ConvT 4x4, stride 2, bn, relu & 256 Conv 3x3, stride 2, bn, lrelu \\
			-----			& 256 Conv 3x3, stride 1, bn, lrelu \\
1 ConvT 4x4, stride 2, sigmoid	& nc Conv 4x4, stride 1, softmax \\		
\hline
\end{tabular} 
\begin{tablenotes}
  \small
  \item *ConvT=transposed convolution 
  \item *relu = rectified linear units
	\item *lrelu=leaky rectified linear units 
	\item *bn=batch normalization
	\item *nc=number of clusters
\end{tablenotes}
\end{threeparttable}
\end{table}

\begin{table}
\centering
\renewcommand{\arraystretch}{1.3}
\begin{threeparttable}
\caption{Clustering error on MNIST dataset for various models}
\label{tab:MNIST_res}

\begin{tabular}{c c }
\hline
Model & Percentage error \\
\hline
DEC (10 clusters)~\cite{xie2016unsupervised}  & 15.6 \\
CatGAN (20 clusters)~\cite{springenberg2015unsupervised} & 4.27\\
Adversarial autoencoder (32 clusters)~\cite{makhzani2015adversarial} & 4.10$\pm$1.13 \\
\hline
Discoder (unregularized) & 24.3$\pm$10.6\\
DisCoder (20 clusters) &  \bf{3.12$\pm$0.93} \\
\hline
\end{tabular} 
\end{threeparttable}
\end{table}

\noindent\textbf{20 newsgroup:} In order to ensure reproducibility, we use a preprocessed version of the dataset\footnote{The pre-processed dataset is available at https://sites.google.com/site/renatocorrea02/textcategorizationdatasets}. We use tf-idf representation for the documents. We train a DisCoder with negative sampling for clustering on this dataset as detailed in Section~\ref{regularization}. The encoding network consists of a single hidden layer with $1000$ units. In order to compute clustering accuracy, we use the same strategy that we had used for MNIST.

The clustering accuracy achieved by the various models on 20 newsgroup dataset is shown in Table~\ref{tab:20news}. For LSA and LDA, we set the number of topics to the number of clusters. We assign each document to the topic with the highest probability for the given document. Again, as in the case of MNIST, a regularized DisCoder outperforms all the other models.

\begin{table}
\centering
\renewcommand{\arraystretch}{1.3}
\begin{threeparttable}
\caption{Clustering accuracy on 20 newsgroup dataset for various models}
\label{tab:20news}

\begin{tabular}{c c}
\hline
Model & Percentage accuracy on 20 newsgroup  \\
\hline
K-means & {40.4} \\
LSA~\cite{landauer1998introduction} & 57.25 \\
LDA~\cite{blei2003latent} & 55.6  \\
\hline
Discoder (unregularized) & 35.11$\pm$8.13 \\
DisCoder  & \bf{ 61.43$\pm$2.24} \\ 
\hline
\end{tabular} 
\end{threeparttable}
\end{table}

\noindent\textbf{CIFAR-10:} Next, we evaluate the performance of the DisCoder on semi-supervised learning task for CIFAR-10 dataset. CIFAR-10 is a relatively complicated image dataset with $50000$ training samples belonging to $10$ classes. There is high variability within each class, making it almost impossible to perform clustering on this dataset. Hence, we use the dataset for semi-supervised classification. We normalize the images to lie between $-1$ and $1$. We use $400$ examples per class as labelled data, and use the rest of the images as unlabelled.

The architecture of the generator and the encoding network used for this task is given in Table~\ref{tab:CIFAR_arch}. The architecture of the encoding network is essentially the same as the architecture of discriminator in Improved GAN~\cite{salimans2016improved}. In particular, the magnitude of weights in all the convolution layers of the encoding network are held fixed during training, and leaky rectified non-linearities with a slope of $.2$ follow every convolution layer. Batch normalization is used in the layers of the generator.

We use feature matching to generate the fake images for our experiments on semi-supervised learning. The classification error achieved by the various models on CIFAR-10 is shown in Table~\ref{tab:CIFARs}. DisCoder achieves the lowest classification error among all the models, significantly outperforming both Imrpoved GAN and CatGAN. 

\begin{figure}[!t]
\centering
\includegraphics[width=2.5in]{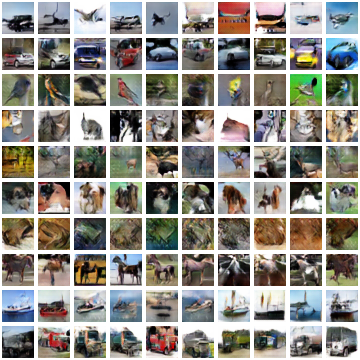}\\
\vspace{.5cm}
\includegraphics[width=2.5in]{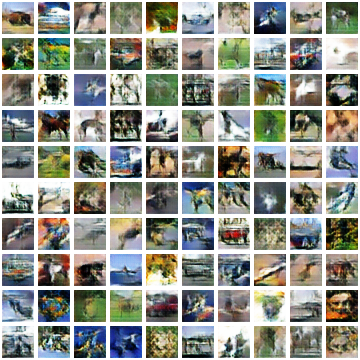}
\caption{(Top) Generated samples from the semi-supervised CIFAR-10 generator using class-conditioned generation. Each row corresponds to a different component of $\fatz$ set to $1$. Since this problem is semi-supervised, $\fatz$ isn't exactly a latent feature. Note that for many classes, the generator is able to preserve the shape of the objects, which is known to be quite challenging for CIFAR-10 images. (Bottom) Samples generated by the generator using feature matching for the same task. Note that the shape of the objects isn't preserved in the images. }
\label{fig_CIFAR}
\end{figure}

We use class-conditioned generation to visualize samples associated with each class learnt by the generator. To generate each row, we set the corresponding component of $\fatz$ to $1$, concatenate it with noise, and pass it through the generator. The corresponding images for the $10$ classes are shown in Figure~\ref{fig_CIFAR}. One can observe that for many classes, the generator is able to capture the shape of the objects quite effectively, which has been known to be quite challenging for CIFAR-10. For comparison, we have shown the images generated by feature matching when trained on CIFAR-10 for the same task. As one can observe, most of the images are meaningless blobs.

\begin{table}
\centering
\renewcommand{\arraystretch}{1.3}
\begin{threeparttable}
\caption{Network architecture for CIFAR10}
\label{tab:CIFAR_arch}
\begin{tabular}{c c }
\hline
Generator & Encoder\\
\hline
512 ConvT 4x4, relu  & dropout(.2) \\
			-----  & 96 Conv 3x3, stride 1, wn, lrelu\\
		 	-----  & 96 Conv 3x3, stride 1, wn, lrelu \\
		 	-----  & 96 Conv 3x3, stride 2, wn, lrelu \\
256 ConvT 4x4, stride 2, bn, relu  &	 dropout(.5)					\\
			-----  & 192 Conv 3x3, stride 1, wn, lrelu\\
		 	-----  & 192 Conv 3x3, stride 1, wn, lrelu \\
		 	-----  & 192 Conv 3x3, stride 2, wn, lrelu \\
128 ConvT 4x4, stride 2, bn, relu & dropout(.5) \\
			-----			& 192 Conv 3x3, stride 1, wn, lrelu \\
			-----			& 192 Conv 1x1, stride 1, wn, lrelu \\
			-----			& 192 Conv 1x1, stride 1, wn, lrelu \\
			-----			& Global average pooling \\
3 ConvT 4x4, stride 2, tanh	& 10 linear,wn, softmax \\		
\hline
\end{tabular} 
\begin{tablenotes}
  \small
  \item *wn= weight normalization
\end{tablenotes}

\end{threeparttable}
\end{table}

\begin{table}
\centering
\renewcommand{\arraystretch}{1.3}
\begin{threeparttable}
\caption{Performance of various models on CIFAR-10 dataset for the task of semi-supervised classification.}
\label{tab:CIFARs}

\begin{tabular}{c c }
\hline
Model & Percentage error using 4000 labeled samples\\
\hline
Ladder network~\cite{rasmus2015semi}  & 20.4$\pm$0.47 \\
CatGAN~\cite{springenberg2015unsupervised} & 19.58$\pm$0.46\\
Improved GAN~\cite{salimans2016improved} & {18.63$\pm$2.32} \\
\hline
Discoder (unregularized) & 31.33$\pm$4.4 \\
DisCoder &  \bf{17.24$\pm$0.43}\\
\hline
\end{tabular} 
\end{threeparttable}
\end{table}

\noindent\textbf{SVHN:} We also evaluate the performance of the DisCoder for semi-supervised learning on 	
 the Street View House Numbers (SVHN) dataset~\cite{netzer2011reading}. The dataset consists of cropped images of digits extracted from house numbers in Google Street View images. As is the case with CIFAR-10, the dataset has high variability, and hence, requires a few labelled examples per class to achieve satisfactory classification result. We use $100$ examples per class as labelled images, while the rest of the images are used as unlabelled images.

\begin{figure}[!t]
\centering
\includegraphics[width=2.5in]{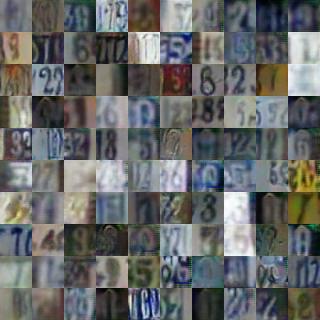}\\
\caption{Samples generated by the generator using feature matching for semi-supervised learning on SVHN.}
\label{fig_SVHN}
\end{figure}

The architecture of the generator and the encoding network used for this task is given in Table~\ref{tab:SVHN_arch}. Again, the architecture of the encoding network is essentially the same as the architecture of discriminator in Improved GAN~\cite{salimans2016improved}. Feature matching is used to generate the fake images for our experiments on semi-supervised learning. The classification error achieved by the various models on SVHN is shown in Table~\ref{tab:SVHN}. As can be observed, regularized DisCoder outperforms Improved GAN by a significant margin, despite using the same network architecture. The samples generated using feature matching are given in Figure~\ref{fig_SVHN}.

\begin{table}
\centering
\renewcommand{\arraystretch}{1.3}
\begin{threeparttable}
\caption{Network architecture for SVHN}
\label{tab:SVHN_arch}
\begin{tabular}{c c }
\hline
Generator & Encoder\\
\hline
512 ConvT 4x4, relu  & dropout(.2) \\
			-----  & 64 Conv 3x3, stride 1, wn, lrelu\\
		 	-----  & 64 Conv 3x3, stride 1, wn, lrelu \\
		 	-----  & 64 Conv 3x3, stride 2, wn, lrelu \\
256 ConvT 4x4, stride 2, bn, relu  &	 dropout(.5)					\\
			-----  & 128 Conv 3x3, stride 1, wn, lrelu\\
		 	-----  & 128 Conv 3x3, stride 1, wn, lrelu \\
		 	-----  & 128 Conv 3x3, stride 2, wn, lrelu \\
128 ConvT 4x4, stride 2, bn, relu & dropout(.5) \\
			-----			& 128 Conv 3x3, stride 1, wn, lrelu \\
			-----			& 128 Conv 1x1, stride 1, wn, lrelu \\
			-----			& 128 Conv 1x1, stride 1, wn, lrelu \\
			-----			& Global average pooling \\
3 ConvT 4x4, stride 2, tanh	& 10 linear,wn, softmax \\		
\hline
\end{tabular} 
\begin{tablenotes}
  \small
  \item *wn= weight normalization
\end{tablenotes}

\end{threeparttable}
\end{table}

\begin{table}
\centering
\renewcommand{\arraystretch}{1.3}
\begin{threeparttable}
\caption{Performance of various models on SVHN dataset for the task of semi-supervised classification.}
\label{tab:SVHN}

\begin{tabular}{c c }
\hline
Model & Percentage error using 4000 labeled samples\\
\hline
SDGM~\cite{maaloe2016auxiliary}  & 16.61$\pm$0.24 \\
ADGM~\cite{maaloe2016auxiliary} & 22.86\\
Improved GAN~\cite{salimans2016improved} & {8.11$\pm$1.13} \\
\hline
Discoder (unregularized) & 16.33$\pm$2.42 \\
DisCoder &  \bf{5.22$\pm$0.12}\\
\hline
\end{tabular} 
\end{threeparttable}
\end{table}

\subsection{Auxiliary classification}
Finally, we train the Discoder to obtain real valued embeddings for the STL-10 dataset. The dataset consists of $100,000$ unlabelled images of size $64 \times 64$ of vehicles and animals. We normalize the images to lie between $-1$ and $1$. The encoding network consists of $3$ convolution layers with $64, 128$ and $256$ filters respectively of filter size $5\times 5$. The last two layers are fully connected layers with $512$ and $10$ neurons respectively. We use weight normalization with fixed norm for all but the last layer.

We use a batch size of $500$ images. Adam optimizer with a learning rate of $.0003$ is used. The encoding network is trained for $800$ epochs. In particular, the objectives in~\eqref{eq:latent_normal} and \eqref{latent_optimize} are optimized alternatively. No adversarial regularization is used for training.
After training, the weights of all but the last layer are frozen. Finally, the last layer is trained using labelled data to minimize classification loss.

The trained network achieves a classificiation accuracy of $71.2$\%  on STL-10 dataset. In contrast, the state of the art on STL-10 using fully supervised training is $70.1$\%~\cite{NIPS2013_5086}.

\section{Conclusion}
In this paper, we introduced a method for encoding the data in an unsupervised manner. We demonstrated the results for the cases when the encoding distribution is either categorical or Gaussian. The learnt embeddings were either used directly in an unsupervised or semi-supervised learning problem or fed to a fully connected neural network for auxiliary classification. Using the learnt embeddings, we were able to achieve the state-of-the-art performance for several well known datasets.

While the objective of an autoencoder encourages it to learn embeddings that minimize reconstruction error, the objective function of a DisCoder encourages the embeddings to be dissimilar to each other. To achieve this task, the DisCoder is forced to capture features that are most discriminative among the samples. For one-hot features, we regularize the model using adversarial regularization to prevent it from overfitting to our initial selection of latent features. However, no such regularization is needed for real valued features because of the continuity of feature space and the stochastic nature of optimization.

In this paper, we have dealt with categorical and Gaussian encoding distributions only. However, the model definition makes no assumptions about the encoding distribution or prior. It will be interesting to explore the utility of the model for handling more complicated embeddings (for instance, a latent Markov chain)  in the future.



\bibliographystyle{IEEEtran}
\bibliography{icdm}
%

\end{document}